\documentclass[11pt]{article}
\usepackage{eamt23}
\usepackage{times}
\usepackage{url}
\usepackage{multirow}
\usepackage{amssymb}

\usepackage{latexsym}
\usepackage{amsmath}
\usepackage{array}
\newcolumntype{$}{>{\global\let\currentrowstyle\relax}}
\newcolumntype{^}{>{\currentrowstyle}}
\newcommand{\rowstyle}[1]{\gdef\currentrowstyle{#1}%
  #1\ignorespaces
}

\usepackage[small,bf]{caption} 
\title{Gender, names and other mysteries: Towards the ambiguous for gender-inclusive translation}

\author{Danielle Saunders\\
  Language Weaver\\
  RWS Group\\
  {\tt dsaunders@rws.com}  \And
  Katrina Olsen\\
  Language Weaver\\
  RWS Group\\
  {\tt kolsen@rws.com}
}
\date{}

\begin{document}
\maketitle
\begin{abstract}
The vast majority of work on gender in MT focuses on `unambiguous' inputs, where gender markers in the source language are expected to be resolved in the output. Conversely, this paper explores the widespread case where the source sentence lacks explicit gender markers, but the target sentence contains them due to richer grammatical gender. We particularly focus on inputs containing  person names.

Investigating such sentence pairs casts a new light on research into MT gender bias and its mitigation. We find that many name-gender co-occurrences in MT data are not resolvable with `unambiguous gender' in the source language, and that  gender-ambiguous examples can make up a large proportion of training examples. From this, we discuss potential steps toward gender-inclusive translation which accepts the ambiguity in both gender and translation.  

\end{abstract}

\section{Introduction}
Different languages express grammatical gender to differing extents. Where  language refers to a person, that person's sociological gender is often expressed via grammatical gender, whether this is simply gendered pronouns in English or profession nouns in German. For machine translation, it is desirable to translate these expressions of gender to the output when they are expressed in the source, and to not express gender in the output which is not implied by the source \cite{savoldi-neutral-2023}.

Gender translation is well-defined when translating into a target language that marks gender in the same way as the source language, for the same parts of speech or a subset of them. 
Difficulties arise when translating into a target language which gender-inflects more parts of speech than the source.
This gender translation challenge can be split into two sub-challenges: unambiguous gender translation and ambiguous gender translation. Unambiguous gender translation generally requires gender markers in the source sentence \cite{renduchintala-williams-2022-investigating}.
The challenge is to resolve gendered terms in the target language consistently with the information in the source sentence. For example, translating `She is an engineer' into German would require that `engineer' be translated with the feminine form, `Ingenieurin', not `Ingenieur'.
Most research to date on gender bias and gender handling in NMT has focused on this unambiguous case \cite{savoldi-etal-2021-gender}.

By contrast, the ambiguous case has no gender markers in the source sentence, either surface-level in the form of gendered parts-of-speech or meta-level in the form of speaker tags or user preference. For example, translating `Taylor is an engineer' into German would still involve choosing a grammatical gender for `engineer', but it is unclear whether the best option is `Ingenieur', `Ingenieurin', or a more inclusive but less commonly used formulation such as `Ingenieur$\ast$in'. The goal here is less well-defined. 
One option is neutralization, avoiding terms implying gender in the output, which may be difficult depending on the output language. 
Another option, used on a small scale in some commercial systems, is annotating and producing multiple output translations where source gender is ambiguous \cite{johnson2018providing}. The latter approach, while UI-dependent, has potential for inclusion of new gender-neutral or other non-binary gendered terms.   

In this paper, we seek to motivate further research into ambiguous-gender translation. We show that ambiguous-gender translation is an underexplored task relative to its prevalence in corpora. We also perform qualitative exploration of the possibilities and challenges of ambiguous gender translation. 

We particularly focus on gender-ambiguous translation with  reference to named entities.
Predicting gender from person name is unreliable and exclusionary (Section \ref{sec:names-really}). However, person names do often co-occur with terms that are gendered in a rich-morphology target language. Sentences containing person names are therefore highly relevant to gender-ambiguous translation.

We first describe related work on gender translation, including some flaws in common treatments of `unambiguous' gender translation. We then describe and use a simple, high-recall method to identify parallel segments meeting our ambiguous gender criterion, where the model must infer gender despite no reliable markers in the source. We analyse the gender characteristics of the results, focusing on English translation into German, French and Spanish in two domains, OpenSubtitles and Europarl. Finally, we describe some possible directions based on our findings towards gender-inclusive translation technologies, with particular reference to inclusivity of those likely to be misgendered by typical name-gender proxies -- anyone whose name does not conform to anglo-centric name-gender associations.

\subsection{Related work}

Various approaches have been taken to gendering named or otherwise ambiguous entities when gender information is available externally, for example by speaker information. Vanmassenhove et al \shortcite{vanmassenhove-etal-2018-getting} incorporate gender information as a tag during training for better translation of first-person sentences. 
Saunders et al \shortcite{saunders-etal-2022-first} rerank n-best translations according to grammatical agreement with a known-gender named entity.

Closer to our approach is work by Wang et al \shortcite{wang-etal-2022-measuring}, which explores the effect of person names on machine translation absent explicit gender information. They take the position that MT \emph{should} use person names as a proxy for gender where no other gender marker is present, and encourage models to treat names assigned to gender categories similarly. Mota et al \shortcite{mota-etal-2022-fast} similarly mask all names with the goal of better name translation, and predict gender from names as male, female or unisex in order to maintain grammatical consistency.  While we also identify inputs lacking explicit gender markers as a key challenge for MT, we differ by treating names as not having a one-to-one mapping with gender, and propose other ways to determine or control target language gender for sentence pairs with person names.

M{\v{e}}chura \shortcite{mechura-2022-taxonomy} is also close to our work in proposing a taxonomy for gender ambiguities in MT inputs, and a schema for resolving them with respect to a target language; their work focuses on professional nouns,  orthogonal to our focus  on named entities.








\section{Unambiguous gender translation?}
\label{sec:unambiguous-really}

In this section, we discuss some assumptions about the resolvability and predictability of gender made in the MT gender literature. 

\subsection{Pronoun coreference is often ambiguous}

Pronouns are often used as gender markers for unambiguous gender. Many machine translation gender test sets including the Gendered Ambiguous Pronoun (GAP) task \cite{webster-etal-2019-gendered}  and WinoMT  \cite{stanovsky-etal-2019-evaluating}  involve performing coreference resolution for one pronoun given more than one entity in a sentence.

Currently, most MT systems operate on individual sentences in isolation. However, language within each sentence cannot be known to only refer to entities within the sentence. While GAP allows a `neither' option for entity resolution, much work in gender translation assumes pronoun coreference is achievable at the sentence-level.

However, pronouns often have multiple plausible antecedents. Even short sentences with a single gendered pronoun and entity (whether person name or profession noun) are not necessarily resolvable. Consider the sentence `The nurse finished his work', used as a debiasing sentence with unambiguous coreference by Saunders and Byrne \shortcite{saunders-byrne-2020-reducing}. As a hypothetical sentence devoid of context, the nurse can be assumed as unambiguously male, since no other entity was mentioned. However, since utterances are created in context, it is equally valid for this sentence to occur after e.g. `The technician left early', meaning that the original `his' coreference is ambiguous.

It is possible to construct truly unambiguous gender-coreferences using, for example, certain types of reflexivization \cite{gonzalez-etal-2020-type,renduchintala-williams-2022-investigating} - and the failure of MT on such unambiguous inputs is an important problem. However, entirely unambiguous inputs are not necessarily common. Instead, as we find in Section \ref{sec-analysis}, even sentences with gendered pronouns are frequently ambiguous.

\subsection{Person names are not reliably gendered}
\label{sec:names-really}

In the NLP literature, names have frequently been used as a proxy for gender \cite{ananya-etal-2019-genderquant,hall-maudslay-etal-2019-name}. Many given names in an anglo-centric context do have a high correlation with referential gender. However, person names in general do not have a one-to-one mapping with person gender. Some given names are differently gendered in different cultural contexts, e.g. `Andrea'. Some people, especially non-binary people, choose names to avoid gendered associations, or choose a name that does not correlate with their referential gender \cite{dev-etal-2021-harms}. Many surnames and nicknames are not clearly gendered -- and gendered given names are frequently also used as surnames, e.g. `James'. 

Specific named entities may have a known gender, but without additional information, resolving those entities can be difficult.  An example is actor Taylor Lautner (he/him), whose wife Taylor (she/her) shares his surname  \cite{lamarelautner}. There may well be further differently gendered Taylor Lautners in the world. 
Some given names and named entities do have a strong correlation with gender in context: `Taylor Lautner' in an entertainment-domain sentence written in the 2010s  is probably referencing the male actor. However, MT systems do not often have access to information about input domain or time-of-writing. Moreover, even if information about a specific individual's gender is available, it may be ethically undesirable to incorporate this information into the system \cite{larson-2017-gender}.

Finally, it is undesirable to assume a certain given name will always correspond to a certain gender, in the same way that it is undesirable to assume a vehicle mechanic is male or a nurse female simply because the vast majority are or historically have been. Indeed, the gender associations of given names change with time \cite{barry1982evolution}, a particular challenge for MT where the input lacks time-of-writing context. For all these reasons, offering multiple gendered outputs to the user or having an human-in-the-loop, controllable machine translation with the ability to define and incorporate user preferences may be required for truly gender-inclusive translation.

\section{Named entities and ambiguous gender translation}
\label{sec:do-ner}

In this section we motivate investigation of named entities for ambiguous gender translation research, and compare named entity recognition (NER) techniques for MT training data.

\subsection{Why named entities?}

While much work on gender in MT has focused on profession nouns, we concentrate instead on named entities for several reasons: connection to a concrete gender, variety, the relative challenge of detection, and anonymity.

\textbf{Concreteness}: In the ambiguous gender case, profession nouns are often, by convention if not inclusively, translated according to the generic masculine \cite{silveira1980generic}. Named entities by contrast are expected to retain their original gender through any coreferent terms that are gendered in the output. Identifying named entities can therefore be a first step to identifying coreferent words which should be gendered consistently with that entity\footnote{Some prefer a mix of differently gendered referential language \cite{dev-etal-2021-harms}. If language can be produced consistently with one gender, producing consistently with a set of genders is a straightforward extension.}. This distinction occurs because person names, unlike profession nouns, are likely to have a concrete referent from the perspective of the input sentence's writer. A person name thus has a referential link to a specific person, whether real or fictional, who has some relationship with gender, even if it is unclear to which person the name refers without the author's context  \cite{kripke1980naming}.

\textbf{Variety}: Names are a far larger and more diverse category compared to profession nouns. While  CareerPlanner\footnote{\url{https://dot-job-descriptions.careerplanner.com}, access Apr 23} lists approximately 12 thousand professions with many near-duplicates, website Forebears\footnote{\url{https://forebears.io/forenames}, access Apr 23} claims 30 million forenames alone. 


\textbf{Challenge:} Detecting person names in natural language presents a different challenge to detecting professional entities. Professional entities are not usually the same in the source and target language and may not be easily distinguishable from other nouns. As a result, identifying professional entities in MT data requires a level of hand curation \cite{prates2019assessing}. By contrast, person names may often be expected to be identical on both source and target. Exceptions such as transliteration of English names e.g. into Chinese characters \cite{wan-verspoor-1998-automatic-english}, or inflection of Slavic named entities \cite{jacquet-etal-2019-jrc}, are more complicated, but usually still predictable. However, in our case of translation from English into German, French and Spanish, we make this simplifying assumption.

\textbf{Anonymity:} Finally, there is an increasing interest in anonymity and privacy with respect to NLP models. Models may therefore also be required to translate appropriately when \emph{a} person name is present but the specific person name or gender information is \emph{not} available. Clearly in this situation it is impossible to know if a translation is `correct' with respect to a specific person's gender---further motivating research and development of systems which expect such ambiguities and handle them gracefully.

\subsection{Method}

\begin{table*}
\centering
\begin{tabular}{$l^l|^c^c^c|^c^c^c|^c^c^c}
\hline
\rowstyle{\bfseries}

& & \% Fr & P ($\uparrow$)& FN ($\downarrow$) & \% De & P ($\uparrow$)& FN ($\downarrow$) & \% Es &  P ($\uparrow$)& FN ($\downarrow$) \\
\hline
\multirow{3}{*}{OS} &Title-copy (TC) & 42.8 & 0.81 & 0.0 & 13.2&  0.80  & 0.03 & 18.6& 0.75& 0.04\\
& Spacy-any (SA) & 30.9 & 0.83&0.13& 9.4 & 0.78& 0.04& 11.8& 0.91 &0.49\\
& Spacy-person (SP) & 19.2& 0.96& 0.34&5.5 & 0.75&0.09& 7.4& 0.97&0.49\\
\hline
\multirow{3}{*}{EP} & Title-copy (TC) & 32.7 & 0.22  & 0.01  &27.0 & 0.33  &  0.01 &19.3 & 0.41& 0.0\\
& Spacy-any (SA) &19.6  & 0.25 & 0.27 & 14.5 &  0.21 &  0.31 & 11.1& 0.46 &0.47\\
 & Spacy-person (SP) & 3.6  & 0.82 & 0.14 &3.7 & 0.81  & 0.26 &3.7 & 0.86&0.35\\

\hline
\end{tabular}
\caption{Percentage of each dataset marked as containing person names using each method in OpenSubtitles (OS) and Europarl (EP) en-\{fr, de, es\}. Column P gives estimated precision of name identification based on human evaluation of 100 randomly sampled sentence pairs marked as containing names. Column FN gives false negative rate estimated likewise on a set of 100 pairs marked as not containing names. 
}
\label{tab:NER}
\end{table*}

\begin{table}
\centering
\begin{tabular}{$ll|^c^c^c^c}
\hline
\rowstyle{\bfseries}
& & Lines (M) &\% N &\% P&\% N$\cap$P \\
\hline
\multirow{3}{*}{OS} & Fr &3.79& 42.8&14.8 & 6.0 \\ 
 & De& 5.85 &13.2 & 11.3& 1.6\\
 & Es &48.30 & 18.6& 14.0& 2.3 \\ 
\hline
\multirow{3}{*}{EP} & Fr & 1.97&32.7 & 3.6& 2.0\\ 
 & De&1.90 & 27.0& 3.6& 2.1\\
 & Es &1.96& 19.3 & 3.6 & 2.4\\ 
\hline

\end{tabular}
\caption{Line counts for OpenSubtitles (OS) and Europarl (EP) datasets used in this paper after preprocessing, and percentages of each containing N - person names identified by TC - or P - binary English pronouns - or both.}
\label{tab-initial-data}
\end{table}

We explore person name detection in parallel data using en-\{fr,de,es\} bitext from OpenSubtitles  \cite{lison-etal-2018-opensubtitles2018} and Europarl \cite{koehn-2005-europarl}. We preprocess the data by removing all exact duplicate sentence pairs and filtering by length ratio and language id. Table \ref{tab-initial-data} gives line counts after preprocessing.

For our `title-copy' (TC) approach, we recognize only names containing Unicode characters matching regex \texttt{[A-Za-zÀ-ž'-\_.]}, beginning with a titlecase character and present in both source and target text. One or more consecutive space-separated tokens can be matched.  While a wide variety of possible exceptions exist \cite{mckenzie-names}, this achieves 100\% recall on a list of 200K person names\footnote{https://github.com/FinNLP/humannames}. 

Additionally, we analyse two techniques using NER with the \texttt{en\_core\_news\_sm} model from Spacy 3.2.3\footnote{https://spacy.io}. `Spacy-any' (SA) refers to the subset of TC entities that are also found using the Spacy model NER pipeline on the  English source sentence. `Spacy-person' (SP) refers to the subset of those named entities that Spacy marks as `person'.  

We conduct human evaluation on randomly-selected 100-sentence samples to estimate the precision and false negative rate for each method. For precision we sample from `detected' sentences, marking if the detected entity is interpreted as a person name given sentence context. For false negative we sample 100 `non-detected' sentences and mark if a person name is present. The two annotators are native English speakers with knowledge of the target languages. German is annotated twice with high inter-rater agreement  (Cohen's $\kappa = 0.945$); French and Spanish are each annotated once.

\subsection{Results: Name detection in parallel data}

For the purposes of this paper, we are primarily interested in recall of person names. This is because our ultimate goal is finding target language that may be associated with any person gender. In Table \ref{tab:NER}  we compare recall in terms of percentage of each dataset retrieved for our three contrastive methods. In terms of recall, we find TC the most effective method, retrieving a far higher percentage of each overall dataset in all cases. However, we do estimate precision and false negative rate,  via human evaluation of a 100-sentence random sample for each metric from the matched sentence pairs. False negative rate tracks retrieved percentage of dataset in most cases.

By human evaluation, we determine the precision of the TC method is quite similar to a generic Spacy NER system for all languages and domains. The Spacy-NER system filtered for person identification has significantly higher precision for French and for the Europarl corpora, but  a correspondingly high false negative rate. Precision is far lower for TC and SA on Europarl: qualitative analysis suggests this is due to a very high proportion of country and organisation names in the Europarl domain.  Even accounting for the lower precision on Europarl,  TC recalls a far larger absolute number of person name examples  compared to SP, suggesting this is a practical alternative for finding person names in MT data. In the remainder of this paper we exclusively use the TC method when selecting lines containing person-names for analysis. 

Given the relative attention in the literature to names versus other gender-associated words such as pronouns, it is interesting to compare their prevalence. In Table \ref{tab-initial-data} we compare the proportion of each dataset containing names -- found using the TC method -- to the proportion with binary pronouns\footnote{\texttt{grep 
-Pwi "(she|her|hers|herself|he|him|\\his|himself)"}.  We use a binary match since we find that `they' is overwhelmingly used in the plural in the OpenSubtitles and Europarl datasets.} on the English side.  In all cases, more names are found, even accounting for  the precisions  determined in Table \ref{tab:NER}. Significantly more segments contain names than pronouns in Europarl and French OpenSubtitles. We also note there is  very little overlap between person names and gendered pronouns. Our findings suggest person names are prevalent enough to be their own gender translation challenge.

\section{Using named entities to identify target gendered language}
\label{sec-analysis}

In this section we use the best-performing TC person name method from Section \ref{sec:do-ner} to extract lines containing likely person-names, and further analyse their characteristics in terms of gendered language and potential coreference.

\subsection{Method}

\begin{table*}
\centering
\begin{tabular}{$ll|^c^c|^c^c|^c^c}
\hline
\rowstyle{\bfseries}

  && \% Fr&  Coref & \% De &  Coref & \% Es& Coref \\
  \hline
\multirow{3}{*}{OS} & Trg-gendered TC  &27.5  & 0.42 &5.1  & 0.42 & 10.9 &0.16\\
& - Subset with no src binary pronouns  &  22.5& 0.29  & 4.2& 0.32 & 9.2 &0.20 \\
& - Subset with  src binary pronouns & 5.0 & 0.43 &0.9 &0.65 & 1.7 &0.27\\
\hline
\multirow{3}{*}{EP} & Trg-gendered TC & 30.2  & 0.16 &19.2  & 0.20& 16.9 &0.36\\
& - Subset with no src binary pronouns   & 28.4  &0.21  & 17.7 &  0.22 &  15.2 & 0.29\\
& - Subset with src binary pronouns    &  1.8 & 0.72  & 1.5 & 0.77& 1.7 &0.72\\
\hline

\end{tabular}
\caption{Percentages of the original datasets for `Title-copy' (TC) lines  containing target language gendered terms (`trg-gendered'). `Coref' is estimated coreference proportion -- labelled if a person name  is coreferent with  gendered language in the target -- based on human evaluation of 100 randomly sampled matching sentence pairs.   Coref scores for the subsets do not necessarily average to the score for the whole set, since evaluation is conducted on independent 100-sentence sample sets.}
\label{tab:gendertarget}
\end{table*}

We perform a dependency parse of each target language sentence to identify the head of each person name, and then subsequently any dependents of that head with masculine or feminine grammatical gender. This produces binary gendered terms likely to be associated with a specific named entity. We use this to find likely ambiguous-gender sentences: those with a named entity on the source side and person-referent gendered  language on the target (referred to as trg-gendered). We do not attempt to filter for other lexically gendered English terms like `mother/father', `fireman/firewoman', etc. We find these terms rare in comparison to names and pronouns.

Note that for our purposes it does not strictly matter whether the named entity is actually coreferent with the gendered terms. This is because we are interested not in  translation but in finding co-occurrences that might trigger gender associations in an MT system, correctly or not. However, we do carry out human evaluation to roughly estimate the proportion of cases where target gender is coreferent with the named entity / pronoun to investigate the questions we raise about gender markers in Section \ref{sec:unambiguous-really}. The same annotation approach is used as in the previous section, with the question now being whether the gendered target language is plausibly coreferent with the person name. Inter-rater agreement across the task for German is slightly lower than for name marking, but still very high (Cohen's $\kappa = 0.888$);

Given parallel data, we identify named entities as in Section \ref{sec:do-ner} using the TC method.  We use the same OpenSubtitles and Europarl data, and Spacy version. For target language parsing we use the relevant \{fr,de,es\} \texttt{core\_news\_sm} model from Spacy. As a bonus, we find  that filtering for possibly-coreferent gendered language on the target side seems to result in a higher precision of named entities that are people.

\subsection{Results: Names vs pronouns with target language gender}
Table \ref{tab:gendertarget} gives counts for lines containing person names, with or without binary gendered pronouns in English using the same pronoun match as the previous section, and likely human-referent gendered terms in the target sentence.

Comparing the TC proportions of each dataset in \ref{tab:gendertarget} to the trg-gendered TC lines of Table \ref{tab:NER}, we find that the proportion of lines with likely person names that also contain binary gendered target language in the same subtree as the name in its dependency parse varies significantly based on language and domain.   For German and Spanish the proportions of TC lines which also have target gendered language is far lower than French, across both OS and EP domains. This may be because these langauges gender fewer parts of speech compared to French.

Source sentences containing likely person names are  unlikely to also contain gendered pronouns even when gendered language was also found on the target side. From Table \ref{tab:gendertarget} over 80\% of sentence pairs with a likely person name alongside target gendered language contained no binary gendered source pronoun. In other words, these sentence pairs are ambiguous-gender inputs with gendered outputs. This supports our hypothesis that person-name inputs are a significant and distinct source of person-gender co-occurrences.  


\subsection{Results: How often are entities coreferent with target gender?}
\begin{table*}
\centering
\begin{tabular}{p{6cm}p{8.5cm}}
\hline
\textbf{Source} & \textbf{Notes}\\
\hline
You let Jax know where she is?		&	Jax could be told Jax's own location, or someone else's location\\
What if Kreski's still doing what he did when his brother was alive? & Both `he' and `his' in the source could refer to the same or different entities, and neither are necessarily Kreski\\
You're asking me to kill my son, Ruth.&Ruth could be the listener or son \\ 
- That was your father, Finn.	&		Finn could be the listener or father\\ 
They slaughtered my wife, Adalind, and my three children.	&	Adalind could be the listener, the wife, or a third victim \\
\hline
\end{tabular}
\caption{Examples of ambiguous English person-name coreference in the en-de OpenSubtitles samples.}
\label{tab:ex}
\end{table*}
Having found sentences containing likely person names, we are interested in answering two questions:

\begin{itemize}
    \item How often is a person name associated with grammatically gendered target language?
    \item How often is a person name actually coreferent with grammatically gendered target language?
\end{itemize}

The first question is answered by the TC-trg-gendered dataset percentages in Table \ref{tab:gendertarget}. The second is addressed by the human evaluated coref proportions. We believe there is an important distinction between these points in terms of what an MT system is likely to learn. 

We note that a model trained to produce gendered target language words with no grounding in the source sentence may well learn inappropriate triggers and potentially exhibit gender bias in the output. Prior research suggests this is the case for professions via the generic masculine \cite{tomalin-etal-2020-rethinking} and for names  \cite{wang-etal-2022-measuring}.  Consider the name `John', which we find is associated with terms in the grammatical masculine vs feminine in a ratio of 2.2:1 in the EngFra OpenSubtitles data. If a system predominantly associates the name with masculine target language, the name may trigger the grammatical masculine even when coreference does not require it. 

While we find that inputs containing person names are frequently associated with gendered language on the target side, actual coreference between names and gendered target language seems rarer. The exception is Europarl sentences containing pronouns, which have a very high proportion of sentences referring to people by both name and title (`president', `commissioner', etc.)

Consistently fewer than a third of the human evaluated lines without pronouns had the person entity likely coreferent with gendered language on the target side. This is further evidence that using person names as a proxy for inflecting target side gender may not be fully reliable. 

Interestingly, lines with person names and binary pronouns on the source side were evaluated as coreferent at a much higher rate. This seems to be because many sentences containing both names and pronouns were ambiguous. During annotation, two primary sources for English coreferent ambiguity were found. One is, as mentioned previously, the possibility of a pronoun referring to an entity not in the sentence. The other is from two interpretations of comma-inserted names---as an appositive modifying the proceeding noun phrase, such as `Thank God my son, Aethelwulf, is alive.' or as a direct address to the listener, such as `Yes, Adam, I'm serious.'\footnote{Both examples found in OpenSubtitles en-de} Examples of both ambiguous cases are given in Table \ref{tab:ex}.


\section{Towards gender-inclusive MT}

In this section, we consider what steps towards gender-inclusive MT we might be able to take with access to a wider range of human-referent gendered data. We consider two broad aspects, training and evaluation.

\subsection{Training gender inclusive systems}

As discussed in related work, prior attempts to keep gender-name consistency in translation have attempted to infer grammatical gender from person name. A more inclusive approach might attempt to infer grammatical gender from target side gender information, and incorporate that information as a tag,  as in Vanmassenhove et al \shortcite{vanmassenhove-etal-2018-getting}  or Saunders et al \shortcite{saunders-etal-2020-neural}. Incorporating a tag would let the model learn to associate gender tags with gendered output in the ambiguous case,  without requiring either external information or names as a proxy for gender.  At inference time, assigning the tag by `name gender' would be equivalent to approaches in the literature, while a more gender-inclusive approach might produce multiple outputs using different gender tags, or allow tags to be user-controlled.

Tagging training data dependent on the gender of target side sentences has further advantages. First, the source gender tag is related directly to the target grammatical gender.  By contrast, leaving a name as a gender proxy assumes a name-gender coreference link which, as found in Table \ref{tab:gendertarget}, is frequently not present. As a second advantage, the process of  tagging these sentence pairs involves directly finding a set of plausibly human-referent gendered language on the target side. These would be a prime target for gender rewriting schemes as described in e.g. Jain et al \shortcite{jain-etal-2021-generating}, potentially including fixed rules to produce neoinflections, but with the bonus of being likely human-referent.

\subsection{Evaluating gender inclusive systems}
 
We have demonstrated a high-recall method for obtaining gender-ambiguous sentence pairs with person names in the source and likely human-referent gendered language in the target. We have also shown that we cannot be confident that the human-referent gendered target language is coreferent with the source side entity. This suggests two potential gender translation evaluation schemas.  

One would evaluate how translations respond to changing input name. This is similar to the approach explored in   Wang et al \shortcite{wang-etal-2022-measuring}, but tests for robustness to name stereotyping instead of name-as-gender proxy. An evaluation scheme on these terms would evaluate -- and lead the way to addressing -- bias affecting those whose names do not exist in an easily predicted `correct' relationship with their referential gender. The aim of such an evaluation scheme would not be to `solve' the gender in each context. Instead a system might control gender in a way related to the \emph{presence} of a person name, but not attempt to ground the gender in that specific name's correlations.

Another possible scheme would evaluate neutralisation. Our method can identify likely human-referent gendered words in a translation hypothesis. If the goal is to avoid groundless gendered words, an MT system could  down-weight hypotheses that contain such words, or an evaluation system score for their presence. While this might not be possible for all target languages, it is increasingly a goal of gender-inclusive translation  \cite{savoldi-neutral-2023}.

\section{Conclusions}
This paper proposes a new perspective on gender-inclusive translation technologies. While most research to date has focused on resolving `unambiguous' gender translation, we discuss the challenges of ambiguous gender translation, where a target language implies gender not grounded in the source. We show that an initial exploration of the ambiguous gender scenario related to named entities suggests possibilities for finding many examples of this scenario in parallel data. We suggest applications for this data both for gender bias mitigation and developing more gender-inclusive systems. Overall, we hope to provide a practical perspective on names, gender, and the inherent ambiguities in gender-inclusive translation.

\bibliography{eamt23}
\bibliographystyle{eamt23}
\end{document}